\documentclass[conference]{IEEEtran}
\IEEEoverridecommandlockouts
\usepackage{cite}
\usepackage{amsmath,amssymb,amsfonts}
\usepackage{algorithmic}
\usepackage{graphicx}
\usepackage{textcomp}
\usepackage{xcolor}

\def\BibTeX{{\rm B\kern-.05em{\sc i\kern-.025em b}\kern-.08em
    T\kern-.1667em\lower.7ex\hbox{E}\kern-.125emX}}
\begin{document}

\title{Gaining Extra Supervision via Multi-task learning for Multi-Modal Video Question Answering
\thanks{* Equally contributed}
}

\author{\IEEEauthorblockN{Junyeong Kim*}
\IEEEauthorblockA{\textit{School of Electrical Engineering} \\
\textit{Korea Advanced Institute of Science and Technology}\\
Daejeon, South Korea \\
junyeong.kim@kaist.ac.kr
\thanks{This research was supported by Samsung Research}}
\and
\IEEEauthorblockN{Minuk Ma*}
\IEEEauthorblockA{\textit{School of Electrical Engineering} \\
\textit{Korea Advanced Institute of Science and Technology}\\
Daejeon, South Korea \\
akalsdnr@kaist.ac.kr}
\and
\IEEEauthorblockN{Kyungsu Kim}
\IEEEauthorblockA{
\textit{Samsung Electronics}\\
Seoul, South Korea \\
ks0326.kim@samsung.com}
\and
\IEEEauthorblockN{Sungjin Kim}
\IEEEauthorblockA{
\textit{Samsung Electronics}\\
Seoul, South Korea \\
sj9373.kim@samsung.com}
\and
\IEEEauthorblockN{Chang D. Yoo}
\IEEEauthorblockA{\textit{School of Electrical Engineering} \\
\textit{Korea Advanced Institute of Science and Technology}\\
Daejeon, South Korea \\
cd\_yoo@kaist.ac.kr}
}

\maketitle

\begin{abstract}
This paper proposes a method to gain extra supervision via multi-task learning for multi-modal video question answering. Multi-modal video question answering is an important task that aims at the joint understanding of vision and language. However, establishing large scale dataset for multi-modal video question answering is expensive and the existing benchmarks are relatively small to provide sufficient supervision. To overcome this challenge, this paper proposes a multi-task learning method which is composed of three main components: (1) multi-modal video question answering network that answers the question based on the both video and subtitle feature, (2) temporal retrieval network that predicts the time in the video clip where the question was generated from and (3) modality alignment network that solves metric learning problem to find correct association of video and subtitle modalities. By simultaneously solving related auxiliary tasks with hierarchically shared intermediate layers, the extra synergistic supervisions are provided. Motivated by curriculum learning, multi-task ratio scheduling is proposed to learn easier task earlier to set inductive bias at the beginning of the training. The experiments on publicly available dataset TVQA shows state-of-the-art results, and ablation studies are conducted to prove the statistical validity. 
\end{abstract}


\section{Introduction} 
\label{sec:1}
Multi-modal video question answering is an important task that aims at the joint understanding of vision and language. 
Its application ranges from multimedia search engine to personal assistant. 
Deep neural network have been successfully applied to various QA tasks including textQA \cite{memory,e2ememory,P18-1160,xiong2018dcn}, imageQA \cite{ASK,SAN,DAN,MUTAN,Anderson2017up-down}, and videoQA \cite{jang-CVPR-2017,xu2017video,motion-appearance,Yu_2018_ECCV} with significant performance improvement. 
Recently, the research on multi-modal videoQA \cite{MovieQA,TVQA,RWMN,FVTA,LMN,MDAM} have also benefited from deep neural networks. 
One of the main challenges of multi-modal video question answering is that the size of existing benchmark datasets (e.g. MovieQA \cite{MovieQA}, PororoQA \cite{PororoQA}, and TVQA \cite{TVQA}) are relatively small to provide sufficient supervision to train QA models on the complex task. 
This paper proposes a method to gain extra supervision via multi-task learning for multi-modal video question answering. 
By solving related auxiliary tasks simultaneously with shared intermediate layers, the model is provided with extra synergistic learning signals and leverages the information from auxiliary tasks to boost question-answering performance.

\begin{figure}[t!]
	\centering
		\includegraphics[width=0.5\textwidth]{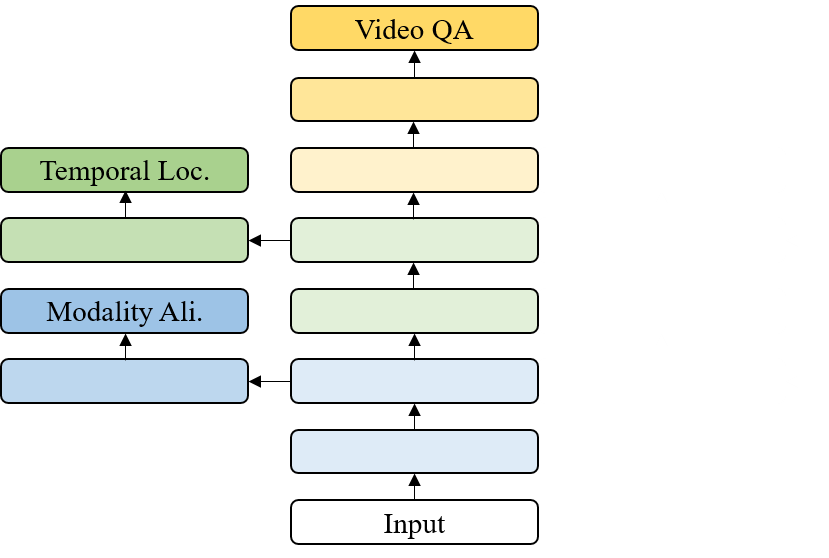}
		\caption{Illustration of the concept of our proposed method for multi-modal video question answering. By simultaneously solving related auxiliary tasks (Modality alignment and Temporal localization) with hierarchically shared intermediate layers, the extra synergistic learning signals are provided to the model. Joint training of all three tasks with proposed multi-task ratio scheduling enables the model to leverage additional information from auxiliary tasks to boost QA performance.}
		\label{fig:task}
\end{figure}

Constructing large-scale dataset for video question answering is difficult. 
The question should be diverse to prevent overfitting to certain question types. 
The correct answer must be correspondent to the question while the other candidate answers should be confusing enough, but not too much misleading or they could be easily excluded. 
The QA pairs generated by human annotators often have a bias. For example, choosing the longest candidate answer gives the accuracy of 25.33\% and 30.41\% for MovieQA \cite{MovieQA} and TVQA \cite{TVQA}, respectively, where the random baseline is 20\%.
Several multi-modal videoQA datasets have been introduced including MovieQA \cite{MovieQA}, PororoQA \cite{PororoQA}, and TVQA \cite{TVQA}. 
However, these benchmark datasets have a relatively small number of QA pairs considering the complexity of the task.
MovieQA \cite{MovieQA} consists of 6,462 QA pairs for the video+subtitles task, PororoQA consists of 8,913 QA pairs, and TVQA consists of 152.5K QA pairs. 
Comparing with imageQA benchmark datasets, VQA v1.0 \cite{VQAv1} dataset consists of 614.2K QA pairs and VQA v2.0 dataset consists of 1.1M QA pairs.

Multi-Task Learning (MTL) is a learning paradigm in machine learning which jointly solves multiple tasks in a single model.
MTL aims to leverage useful information contained in multiple related tasks to gain positive synergies across all the tasks.
For example, tasks like temporal localization and visual-semantic alignment have been found useful for each other when trained jointly \cite{tall}. 
To solve multi-modal videoQA, this paper proceeds by analogy to human intelligence. 
Humans would first have to know the proper association between vision and language. 
On top of that, humans would attempt to localize the moment which is relevant to answering the question. 
Finally, humans could learn how to answer the questions. 
We formulated this as a multi-task learning problem and designed two auxiliary tasks.

This paper proposes a method to gain extra supervision via multi-task learning for multi-modal videoQA. 
Solving auxiliary tasks simultaneously with the QA task can provide synergistic learning signals. 
On top of the QA network based on Jie et al.\cite{TVQA}, we introduce two auxiliary tasks that hierarchically share parameters with the QA network as depicted in Fig. \ref{fig:task}. 
One auxiliary task is modality alignment which aims at correctly associating video and subtitle features. 
It shares parameters with the lower layers of the QA network. 
The other task is temporal localization which aims at finding the moment in the video clip that is most relevant to answering the current question. 
It shares parameters with the higher layers of the QA network. 
In order to control the timing and strength of the objective of each task, multi-task ratio scheduling is proposed.
Motivated by curriculum learning \cite{curriculum}, multi-task ratio scheduling attempts to learn easier task earlier to set inductive bias at the beginning of the training.
The main contribution of this paper is summarized as follows.
(1) Multi-task learning method for multi-modal videoQA is proposed which recorded state-of-the-art performance on TVQA benchmark. 
(2) Multi-task ratio scheduling is proposed to efficiently reflect the objectives of each task during training.

The rest of this paper is organized as follows. 
First, Sec.\ref{sec:related} describes the related works on this paper. 
Then, the proposed method is elaborated in Sec.\ref{sec:proposed}. 
Sec.\ref{sec:exp} describes dataset and experimental results. 
Finally, Sec.\ref{sec:conclude} concludes the paper.

\section{Related work} 
\label{sec:related}

In this section, we introduce the related works of our paper in four categories; multi-task learning, modality alignment, temporal localization and multi-modal video question answering.

\subsection{Multi-Task Learning}
Multi-task learning aims at jointly solving multiple related tasks with a single model. 
By sharing parameters across related tasks, the model can generalize better on the original task. 
Most of the multi-task learning methods share the hidden layers across all tasks and have task-specific output layers for each task. 
Starting from the work of Kaufmann et al.\cite{Caruana93multitasklearning}, there have been rich research on multi-task learning across the majority of applications of machine learning from computer vision (CV) \cite{fastrcnn} to natural language processing (NLP) \cite{Collobert:2008:UAN:1390156.1390177}. 
Kim et al.\cite{JYKIM} proposed Deep Partial Person Re-identification (DPPR) that jointly learns person classification and person re-identification for partial person re-identification. 
Object detection architectures such as Fast R-CNN \cite{fastrcnn} and Faster R-CNN \cite{fasterrcnn} used multi-task loss for bounding box regression and object classification. 
Tiao et al.\cite{xiaoli2017joint} tackled the task of Person Search \cite{xiaoli2017joint} by jointly learning pedestrian detection and person re-identification. 
Recently, Li et el.\cite{iqan} proposed the invertible Question Answering Network (iQAN) to leverage the complementary relations between questions and answers in the image by jointly learning the Visual Question Answering (VQA) and Visual Question Generation (VQG) tasks.

\subsection{Modality Alignment}
As an auxiliary task of our proposed method, we jointly learn the modality alignment and the temporal localization along with the multi-modal video question answering. 
Both tasks have been extensively studied in the field of deep learning.
Karpathy et al. \cite{Karpathy} proposed a method that captures the inter-modal correspondences between vision and language to generate natural language description of the given image. 
The latent alignment between the segments of the sentence and the region of the image is learned with a structured max-margin loss. 
Castrej{\'{o}}n et al. \cite{crossmodal} proposed a method that learn cross-modal scene representations that transfer across modalities. 
By regularizing cross-modal CNNs to have shared representation, the resulting representation is agnostic of the modality.
Yu et al. \cite{jsfusion} proposed Joint Sequence Fusion (JSFusion) model that can measure semantic similarity between any pairs of multimodal sequence data.
Hierarchical attention mechanism is utilized to learn matching representation patterns among modalities.

\subsection{Temporal Localization}
Temporal localization aims at localizing temporal parts from the given video.
Hendricks et al.\cite{LocalizingMoments} proposed the Moment Context Network (MCN) for temporal localization with natural language query. 
The MCN effectively localizes temporal parts related to natural language query by integrating local and global video feature over time.
Gao et al.\cite{tall} proposed a multi-task learning approach for temporal localization with natural language query. 
Location regression and visual-semantic alignment are jointly learned for temporal localization.
Temporal Unit Regression Network (TURN) \cite{TURN} jointly predicts action proposals and refines the temporal boundaries by temporal coordinate regression. 
Long untrimmed video is decomposed into video clips, which are reused as basic building blocks of temporal proposals for fast computation. 

\begin{figure*}
	\includegraphics[width=\textwidth,height=7cm]{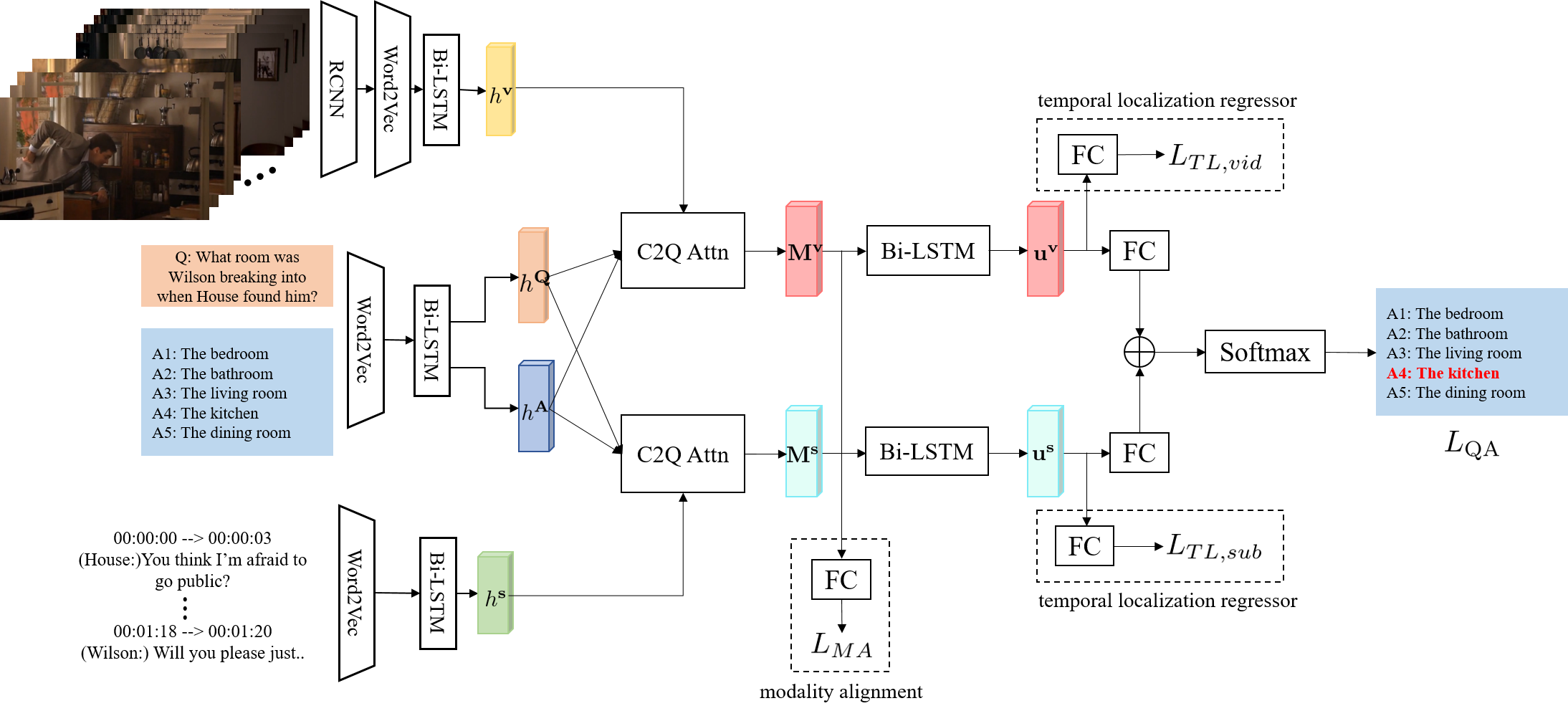}
	\caption{The overall architecture of proposed method for multi-modal video question answering. The proposed network is composed of three networks which share intermediate layers; QA network, modality alignment network, and temporal localization network.}
	\label{fig:overall}
\end{figure*}

\subsection{Multi-Modal Video Question Answering}
Recently, the research on multi-modal videoQA \cite{MovieQA,TVQA,RWMN,FVTA,LMN,MDAM} leverages additional text modality such as subtitle along with video modality for the joint understanding of vision and language. 
There are various benchmark datasets for multi-modal videoQA including MovieQA \cite{MovieQA}, PororoQA \cite{PororoQA} and TVQA \cite{TVQA}. 
Multi-modal videoQA is a challenging task for its relatively small size of benchmark datasets.
The majority of the methods on multi-modal videoQA are motivated by memory-augmented architecture \cite{e2ememory}. 
Tapaswi et al. \cite{MovieQA} utilized memory network (MemN2N) \cite{e2ememory} to store video clips into memory and retrieve required information for answering question. 
Read-Write Memory Network (RWMN) \cite{RWMN} replaces the fully-connected layers in memory network \cite{e2ememory} into the convolutional layers to capture local information in each memory slot. 
After the video and subtitle features are fused using bilinear operation, convolutional write/read networks store/retrieve information, respectively.
Focal Visual-Text Attention (FVTA) \cite{FVTA} applied hierarchical attention mechanism on three-dimensional tensor of question, video and text to dynamically determine which modality and what time to attend for question answering. 
Multimodal Dual Attention Memory (MDAM) \cite{MDAM} applied multi-head attention mechanism \cite{attn} to learn the latent representation of multi-modal inputs.

\section{The proposed method} 
\label{sec:proposed}

In this section, we describe the proposed method and training procedure in details. 
Fig. \ref{fig:overall} gives the overview of overall architecture which fully utilizes multi-modal inputs (video and subtitle) and QA pairs to answer the question. 
The proposed method is composed of three networks which hierarchically shares intermediate layers; question-answering (QA) network, modality alignment network and temporal localization network. 
Note that we utilized QA network proposed by Jie et al. \cite{TVQA}. 

\subsection{Problem Formulation}
\label{ssec:formulation}

The formal definition of Multi-modal Video Question Answering (QA) is as follows. The inputs to the model are (1) a video clip $\mathbf{v}$, (2) a subtitle corresponding to the video clip $\mathbf{s}$, (3) a question $\mathbf{q}$ and (4) five candidate answers $\mathbf{a} = \{a_i\}_{i=0}^4$ where only one is correct. The task is to predict correct answer for the question and the objective of training is to learn model parameters to maximize the following log-likelihood:
\begin{equation}
\hat{\theta} = \text{arg}\underset{\theta}{\text{max}}\sum_\mathcal{D} \log P(\textbf{y}|\textbf{v}, \textbf{s}, \textbf{q}, \textbf{a};\theta),
\end{equation}
where $\mathcal{D}$ denotes the dataset, $\theta$ represents the model parameters and $\textbf{y}$ denotes the correct answer out of five candidates.

\subsection{Feature Extraction}
\label{ssec:feat}

Before describing the proposed method, we first introduce the feature extraction method. For a fair comparison, we followed the same video and text feature extraction procedure used in previous work \cite{TVQA} and fixed them during training.

\subsubsection{Video Features}
We extracted two types of video features; ImageNet feature and visual concept feature. First, the frames are extracted from each video clip $\textbf{v}$ with the rate of 3fps.

\textbf{ImageNet Feature}: For each frame, the feature vector of size 2048D is extracted from \textquotedblleft Average Pooling\textquotedblright\ layer of ResNet-101 \cite{resnet} trained on ImageNet Benchmark \cite{imagenet}. The frame feature corresponding to the same video clip are first L2-normalized and then stacked, forming ImageNet feature $\mathbf{V}^{img} \in \mathbb{R}^{n_{img} \times 2048}$, where $n_{img}$ represents the number of frames extracted from the video clip.

\textbf{Visual Concept Feature}: Inspired by recent works \cite{DBLP:conf/emnlp/YinO17,TVQA} that use detected object labels as visual inputs instead of using CNN features directly, we also extracted detected labels which are referred to as visual concept feature \cite{TVQA}. Faster R-CNN \cite{fasterrcnn} trained on Visual Genome Benchmark \cite{Krishna2017} is utilized to detect objects in each frame. After collecting every detected concept for each video clip over all of the frames and eliminating the overlapping concepts, we utilized GloVe \cite{glove} to embed each concept into feature representation. The resulting visual concept feature is represented as $\mathbf{V}^{cpt} \in \mathbb{R}^{n_{cpt} \times 300}$, where $n_{cpt}$ denotes the number of concepts in video clip.

\subsubsection{Text Features}
We used GloVe \cite{glove} to embed words into feature representations. Every sentences in each subtitle $\mathbf{S}$ are flattened and tokenized into a sequence of words and GloVe \cite{glove} embeds the sequence of words into subtitle feature denoted as $\mathbf{S}^{rep} \in \mathbb{R}^{n_\mathbf{S} \times 300}$, where $n_{\mathbf{S}}$ represents the number of words in the subtitle. Question feature $\mathbf{q}^{rep} \in \mathbb{R}^{n_{\mathbf{q}} \times 300}$ and candidate answer feature $a_i^{rep} \in \mathbb{R}^{n_{a_i} \times 300}$ are also embedded similarly, where $n_\mathbf{q}$ and $n_{a_i}$ are the number of words in question $\mathbf{q}$ and candidate answer $a_i$, respectively.

\subsection{Question-Answering Network}
\label{ssec:qa}

Now, we describe the QA network proposed by Jie et al. \cite{TVQA}. First, bi-directional LSTM (bi-LSTM) is used to encode the both video and text features into embedding space. The bi-LSTM consists of two LSTM layers; forward LSTM $\overset{\rightarrow}{f}$ and backward LSTM $\overset{\leftarrow}{f}$. For the input sequence of $\{x_i\}_{t=1}^T$, the forward LSTM $\overset{\rightarrow}{f}$ encodes the input sequence in forward order (from $x_1$ to $x_T$) into hidden states ($\overset{\rightarrow}{h}_1, \cdots, \overset{\rightarrow}{h}_T$). The backward LSTM $\overset{\leftarrow}{f}$ encodes the input sequence in backward order (from $x_T$ to $x_1$) and generate hidden states ($\overset{\leftarrow}{h}_1, \cdots, \overset{\leftarrow}{h}_T$). The hidden states from both directions at each time step are stacked to obtain resulting hidden representation, i.e. $h_t = [\overset{\rightarrow}{h}_i; \overset{\leftarrow}{h}_i]$ where $[;]$ represents concatenation. Subtitle $\mathbf{S}^{rep}$, question $\mathbf{q}^{rep}$, candidate answer $a_i^{rep}$ and visual concept $\mathbf{V}^{cpt}$ features are now encoded by bi-LSTM and denoted as $h^{\mathbf{S}} \in \mathbb{R}^{n_\mathbf{S} \times 2d}, h^{\mathbf{q}} \in \mathbb{R}^{n_\mathbf{q} \times 2d}, h^{a_i} \in \mathbb{R}^{n_{a_i} \times 2d}$ and $h^{cpt} \in \mathbb{R}^{n_{cpt} \times 2d}$, respectively. Here, $d$ is the size of hidden state which is set to $150$ in this experiment. Similarly for the ImageNet feature $\mathbf{V}^{img}$, it is first fed into fully-connected layer with $tanh$ activation function to project into word space, then encoded by Bi-LSTM producing $h^{img} \in \mathbb{R}^{n_{img} \times 2d}$.

The context-query attention layer \cite{biattn1,biattn2} is utilized to jointly model the encoded context (e.g. video, subtitle) and query (e.g. question, candidate answers). It feeds a set of context vectors $\mathbf{C} \in \mathbb{R}^{n \times 2d}$ and a set of query vectors $\mathbf{Q} \in \mathbb{R}^{m \times 2d}$ as inputs, and constructs context-to-query attention matrix $\mathbf{A}$. The context-to-query attention is generated as follows: First, the similarities between each pair of context vector and query vector are computed, producing a similarity matrix $\mathbf{S} \in \mathbb{R}^{n \times m}$, where $\mathbf{S}_{ij}$ represents the similarity between $i$-th context word $\mathbf{C}_i$ and $j$-th query word $\mathbf{Q}_j$. Instead of the original trilinear function \cite{biattn1}, dot product is utilized to calculate similarity, i.e. $\mathbf{S}_{ij} = \mathbf{C}_i \cdot \mathbf{Q}_j$. Then, we normalize each row of similarity matrix by applying the softmax function, producing a matrix $\bar{\mathbf{S}}$. Finally, context-to-query attention which contains the attended query vectors for the entire context is computed as $\mathbf{A} = \bar{\mathbf{S}}\mathbf{Q} \in \mathbb{R}^{n \times 2d}$. The context-to-query attention signifies which word in query is most relevant to the each word in context.

Consider the upper stream of the Fig.\ref{fig:overall} where the visual concept is used as the context for context-query attention layer. The question and candidate answer are considered as the query to generate the context-to-query attentions $\mathbf{A}^{cpt,q}, \mathbf{A}^{cpt,a_i} \in \mathbb{R}^{n_{cpt} \times 2d}$, respectively. The context-to-query attentions are then fused with context as follows:
\begin{equation}
\mathbf{M}^{cpt, a_i} = [H^{cpt};A^{cpt,q};A^{cpt,a_i}; \\ H^{cpt}\odot A^{cpt,q};H^{cpt}\odot A^{cpt,a_i}],
\end{equation} 
where $\odot$ denotes element-wise multiplication.

Finally, the fused feature vector $\mathbf{M}^{cpt, a_i} \in \mathbb{R}^{n_{cpt} \times 10d}$ is again fed into Bi-LSTM and max-pooled along time to get final feature vector $\mathbf{u}^{cpt, a_i} \in \mathbb{R}^{10d}$ for each candidate answer $a_i$. The prediction score is obtained by applying linear fully-connected layer on a set of final feature vectors $\{\mathbf{u}^{cpt, a_i}\}_{i=1}^{5}$. Prediction score for bottom stream can be computed similarly by utilizing the subtitle as the context for context-query attention layer. The prediction score for each stream is summed to get final score and $softmax$ function is applied to produce answer probability $\hat{\mathbf{y}}$. The cross-entropy loss function is used to train QA model:
\begin{equation}
L_{QA} = -\sum_{i=1}^{5}\mathbf{y}_i \log \hat{\mathbf{y}}_i + (1 - \mathbf{y}_i)\log (1 - \hat{\mathbf{y}_i}).
\end{equation}

\begin{figure}[t]
	\centering
	\includegraphics[width=\linewidth]{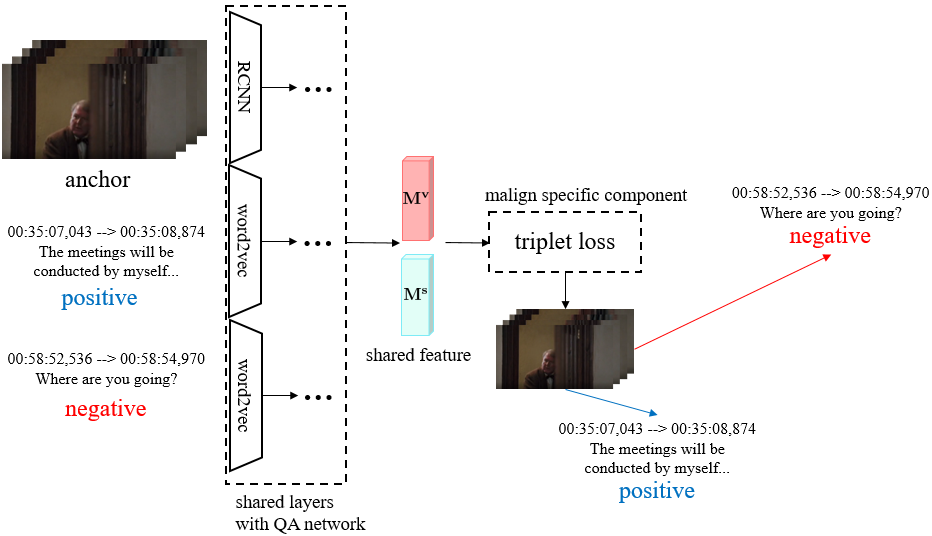}
	\caption{Illustration of modality alignment network. Modality alignment solves multi-modal metric learning problem where it attempts to find correct association of video and subtitle.  }
	\label{fig:malign}
\end{figure}

\subsection{Modality Alignment Network}
\label{ssec:malign}
Modality alignment network regards the pairing of video and subtitle as supervision and attempts to predict the correct alignment between two modalities. 
It shares parameters with the lower layers of the QA network.
After the video and subtitles are embedded into the common space forming $h^{cpt}$ and $h^{\mathbf{S}}$, we denote $\{h^{cpt}_i, h^{\mathbf{S}}_i\}$ as the positive pair and $\{h^{cpt}_i, h^{\mathbf{S}}_{i^-}\}$ as the negative pair of encoded video-subtitle features where $i$ represents the index of element in each mini-batch and $i^-$ represents the index of elements in each mini-batch except $i$.
Each training mini-batch, composed of total $B$ video-subtitle pairs, includes single positive pair and $B-1$ negative samples.

The objective of modality alignment network is to make the features of positive pair get closer in the embedding space, and the negative pairs get farther. 
Intuitively, the video-subtitle pair should have a high matching score if its words have a confident support in the video.
We formulated this as a metric learning problem. 
Motivated by Hoffer et al. \cite{triplet}, we utilized max-margin loss to pull features of positive pairs and push features of negative pairs. 
The distance between positive pair is defined as $d^+ = ||h^{cpt}_i - h^{\mathbf{S}}_i||_2$ and the distance between negative pair is defined as $d^- = ||h^{cpt}_i - h^{\mathbf{S}}_{i^-}||_2$ where $||\cdot||_2$ denotes the $l^2$-norm of vector. 
The modality alignment loss constrains the positive distance $d^+$ to be smaller than $d^-$ by the margin $\tau$:
\begin{equation}
L_{MA} = \max(0, \tau - d^- + d^+).
\end{equation}

\begin{figure}[t]
	\centering
	\includegraphics[width=\linewidth]{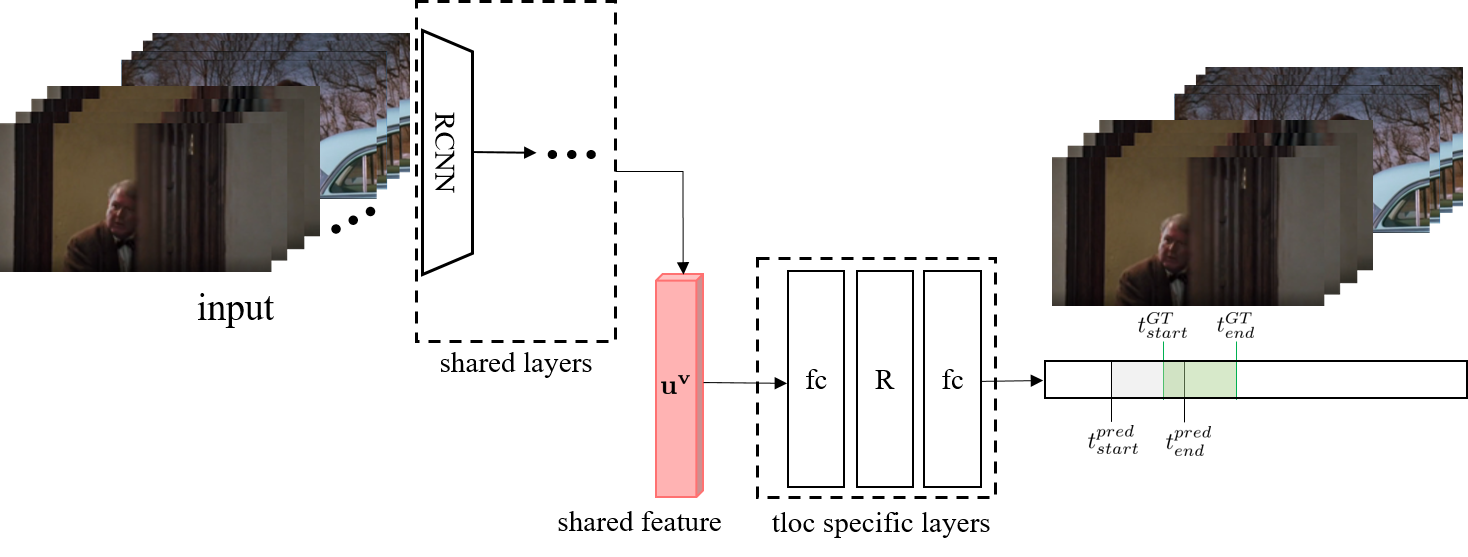}
	\caption{Illustration of temporal localization network. For simplicity, only upper stream that process video input is drawn. Temporal localization network solves regression problem to predict the start and end time of ground truth moment where the question was generated from. }
	\label{fig:tloc}
\end{figure}

\subsection{Temporal Localization Network}
\label{ssec:tloc}
Temporal localization network localizes the temporal part relevant to the question. The moment in the clips where the question was generated from is regarded as supervision. 
It shares parameters with the higher layers of the QA network.
We formulate the objective of temporal localization network as a regression problem.
For each stream, the final feature vectors $\{\mathbf{u}^{a_i} \}_{i=1}^{5}$ are concatenated and used to regress the start point ($t_{start}^{GT}$) and end point ($t_{end}^{GT}$) of ground-truth moment of question generation. 
We normalized the $t_{start}^{GT}, t_{end}^{GT}$ by the length of the video clip to have the value between 0 and 1.
The loss function for the temporal localization network contains two terms:
\begin{equation}
L_{TL} = L_{reg} - L_{overlap}.
\end{equation}
The first term is straightforward regression loss which is the mean squared error between ground-truth and prediction. 
The second term is referred to as \textit{overlap loss} which considers the overlap between the ground-truth and prediction. 
The two terms are formulated as follows:

\begin{eqnarray}
L_{reg} &=& ||t^{GT} - t^{pred}||_2,\\
L_{overlap}&=& \frac{O(t^{GT}, t^{pred})}{t_{end}^{GT} - t_{start}^{GT}}.
\end{eqnarray}
where $O(t^{GT}, t^{pred})$ represents the length of overlap between $t^{GT}$ and $t^{pred}$ which can be formulated as $O(t^{GT}, t^{pred}) = \max(0,\min(t_{end}^{GT},t_{end}^{GT})-\max(t_{start}^{GT}-t_{start}^{pred}))$.

\subsection{Multi-Task Ratio Scheduling}
\label{ssec:trainig}
The entire network is trained by simultaneously optimizing the aforementioned three loss functions; one for QA network, modality alignment network, and temporal localization network, respectively.
The total loss to be minimized during training is the weighted sum of all three losses:
\begin{equation}
L = \alpha_{QA} L_{QA} + \alpha_{MA} L_{MA} + \alpha_{TL} L_{TL}
\end{equation}
where $\alpha_{QA}$, $\alpha_{MA}$, and $\alpha_{TL}$ are the weights for each loss function. 
In order to control the timing and strength of the objective of each task, multi-task ratio scheduling is proposed to schedule the weights for each loss function.
Motivated by curriculum learning \cite{curriculum}, simple tasks are focused more at the early stage of training and complex tasks are focused later.
Among the designed tasks, the task of modality alignment and temporal localization are easier than question-answering.
Therefore initially, the weight for modality alignment $\alpha_{MA}$ is set higher than the other weights to facilitate solving modality alignment task.
Then, the weight for temporal localization $\alpha_{TL}$ is set higher and finally, the weight for question-answering $\alpha_{QA}$ is set to the highest to solve the multi-modal video question-answering task. 

\section{Experiments} 
\label{sec:exp}
This section provides the experimental details and results of our proposed method. 
First, the benchmark dataset used to train and evaluate the proposed model is introduced. 
Then, we describe the experimental details. 
Finally, we provide quantitative results with ablation study.

\subsection{Dataset}
\label{ssec:dataset}
The TVQA Benchmark \cite{TVQA} is a multi-modal video question answering dataset. 
It is collected on 6 long-running TV shows from 3 genres: (1) sitcoms: \textit{The Big Bang Theory, How I Met Your Mother, Friends}, (2) medical dramas: \textit{Grey's Anatomy, House} and (3) crime drama: \textit{Castle}. 
Total 21,793 short clips of 60/90 seconds are segmented for TVQA \cite{TVQA}, accompanied with corresponding subtitles and character names.
The questions in TVQA Benchmark is composed of following formal \textquotedblleft [What/How/Where/Why/...] \underline{\ \ \ \ \ \ \ \ \ } [when/before/after] \underline{\ \ \ \ \ \ \ \ \ }?\textquotedblright\ where second part localizes required point within the video clip for answering the question and first part provides the question on that point.
The overall number of multiple-choice question-answer pairs are 152.5k, where train split contains 122,039 QA pairs, validation split contains 15,252 QA pairs and test split contains 7,623 QA pairs.
Each QA pair has five candidate answers, but only one of them is correct.
The performance of each model is measured by multiple-choice question answering accuracy.

\subsection{Implementation Details}
\label{ssec:implementation}
The proposed method was implemented using PyTorch framework. All of the experiments in this paper were performed under CUDA acceleration with single NVIDIA TITAN Xp (12GB of memory) GPU and trained using the Adam optimizer \cite{adam} with the learning rate of 0.0003 and mini-batch size of 32. On average, it took almost 12 hours for our proposed model to converge.

\subsection{Experimental Results} 

\begin{table}[t]
	\caption{Accuracy comparison on the validation and test set of TVQA benchmark. The symbol meanings are Q=Question, S=Subtitle, V=Video, img=ImageNet features, reg=regional visual features, cpt=visual concept features. Our method achieves the state-of-the-art performance. The test set accuracy of our proposed method is obtained from online evaluation server. The symbol \textquoteleft-\textquoteright\ indicates that the performance is not provided.}
	\label{tab:acc}
	\begin{center}
		\begin{tabular}{l||c|c|c}
			\hline
			Methods                    & Video  Feature & valid Acc. & test Acc.    \\
			\hline \hline
			Random                  & - & -  & 20.00                 \\
			Longest Answer                   & - & - & 30.41                 \\
			\hline
			TVQA S+Q \cite{TVQA}					& - & - & 63.14 \\
			\hline
			TVQA V+Q \cite{TVQA}                    & img & - & 42.67 \\
			TVQA V+Q \cite{TVQA}                    & reg & - & 42.75 \\
			TVQA V+Q \cite{TVQA}                    & cpt & - & 43.38 \\
			\hline
			TVQA S+V+Q \cite{TVQA}					& img & - & 63.57 \\
			TVQA S+V+Q \cite{TVQA}					& reg & - & 63.19 \\
			TVQA S+V+Q \cite{TVQA}					& cpt & - & 65.46 \\
			\hline \hline
			ours S+Q                 & - & 64.36 & 64.63                 \\
			\hline
			ours V+Q                 & img & 42.13 & 42.79 \\
			ours V+Q                 & cpt & 43.45 & 44.42 \\
			\hline
			ours S+V+Q				 & img & 63.99 & 64.53 \\
			ours S+V+Q				 & cpt & \bf{66.22} & \bf{67.05} \\
			\hline
		\end{tabular}
	\end{center}
	
\end{table}

The experimental results are summarized in table \ref{tab:acc}. 
We compared the performance of our proposed method with the results reported in the TVQA paper \cite{TVQA}. 
The random baseline shows 20.00\% test accuracy for the task of multiple-choice question-answering with 5 candidate answers. 
The longest answer baseline selects the longest answer for each question. 
It achieves the performance of 30.41\% which indicates that the correct answers tend to be longer than the wrong answers.
Note that the validation accuracy of TVQA methods is not reported in the original paper \cite{TVQA}.

Our subtitle-only (ours S+Q) method achieves test accuracy of 64.63\% which is 34.22\% higher than the longest answer baseline and 1.49\% higher than the subtitle-only TVQA baseline (TVQA S+Q). 
Our video-only (ours V+Q) methods achieve the performance of 42.79\% and 44.42\% for ImageNet and visual concept feature, respectively.
Compared to the TVQA baseline, our results on video-only methods obtains 0.12\% and 1.04\% performance boost.
For our uni-model results (ours S+Q and ours V+Q), only temporal localization loss is utilized as an auxiliary loss. 
Our full model (ours S+V+Q) with ImageNet feature achieves the performance of 64.53\% which is 0.96\% higher than the TVQA S+V+Q with ImageNet feature. 
Our full (ours S+V+Q) method with visual concept feature achieves the state-of-the-art result on TVQA dataset with the performance of 67.05\% which is 1.59\% higher accuracy than the runner-up model, TVQA S+V+Q with visual concept feature. 
The experimental results verify that multi-task learning can bring additional performance boost especially when the task is complex.
Especially, our S+Q model, which only uses subtitle and question but not video, achieved higher performance (64.63\%) than TVQA S+V+Q model with ImageNet feature (63.57\%). 
This shows competent performance improvement caused by extra supervision from multi-task learning. 

\begin{table}[t]
	\caption{Result of ablation study between model variants. Both the video and subtitles are utilized in this study. The use of temporal localization loss, modality alignment loss are denoted as \textquoteleft+ TL\textquoteright\ and \textquoteleft+ MA\textquoteright. The last column (denoted as $\Delta$) shows the performance drop compared to the full model.}
	\label{tab:abl}
	\begin{center}
		\begin{tabular}{l||c|c|c}
			\hline
			Methods                 & Video Feature & valid Acc.  & $\Delta$ \\
			\hline \hline
			QA                 		& img & 63.14 & -0.85\\
			QA + MA                 & img & 63.67 & -0.32\\
			QA + TL					& img & 63.49 & -0.50\\
			QA + MA + TL 			& img & 63.99 & -\\
			\hline
			QA						& cpt & 65.03 & -1.19\\			
			QA + MA					& cpt & 65.79 & -0.43\\
			QA + TL					& cpt & 65.64 & -0.58\\
			QA + MA + TL			& cpt & 66.22 & -\\
			\hline
		\end{tabular}
	\end{center}
	
\end{table}

For ablation study, we only report the validation accuracy since test accuracy can only be measured through an online evaluation server for finite times. 
Table \ref{tab:abl} summarizes the results of ablation study. 
Overall, the visual concept feature gives higher performance than ImageNet feature as reported in \cite{TVQA}. 
The first block of Table \ref{tab:abl} shows the ablation study with the ImageNet feature and the second block of Table \ref{tab:abl} shows the ablation study with the visual concept feature.
Multi-task learning with temporal localization and modality alignment showed a meaningful increase in performance for both cases when used ImageNet and visual concept features. 
The ablation study suggests that solving auxiliary tasks together with the main task of video question answering can improve performance. 
Modality alignment brought higher performance gain than temporal localization. 
The amount of performance gain may differ by implementation detail though, the performance gain can differ by the choice and the scheduling of auxiliary tasks. 
The more relevant and helpful the auxiliary task is to the main task, the higher the performance gain is.

\section{Conclusion}
\label{sec:conclude}

In this paper, we proposed a method to gain extra supervision via multi-task learning for multi-modal video question answering. We argue that the existing benchmark datasets on multi-modal video question answering are relatively small to provide sufficient supervision. To overcome this challenge, this paper proposes a multi-task learning method which is composed of three main components: (1) multi-modal video question answering network that answers the question based on the both video and subtitle feature, (2) temporal retrieval network that predicts the time in the video clip where the question was generated from and (3) modality alignment network that solves metric learning problem to find correct association of video and subtitle modalities. Motivated by curriculum learning, multi-task ratio scheduling is proposed to learn easier task earlier to set inductive bias at the beginning of the training. The experiments on publicly available dataset TVQA shows state-of-the-art results, and ablation studies are conducted to prove the statistical validity. 
\bibliographystyle{IEEEtran}
\bibliography{ijcnn}

\end{document}